\documentclass[12pt,authoryear,nopreprintline]{elsarticle}

\usepackage[outdir=images/lhf/]{epstopdf}

\usepackage{graphicx}%
\usepackage{graphics}%
\usepackage{multirow}%
\usepackage{amsmath,amssymb,amsfonts}%
\usepackage{stmaryrd}
\usepackage{amsthm}%
\usepackage{mathrsfs}%
\usepackage[title]{appendix}%
\usepackage{xcolor}%
\usepackage{textcomp}%
\usepackage{manyfoot}%
\usepackage{booktabs}%
\usepackage{algorithm}%
\usepackage{algorithmicx}%
\usepackage{algpseudocode}%
\usepackage{listings}%
\usepackage{adjustbox}%
\usepackage[justification=justified, format=plain]{caption}
\usepackage{hyperref}
\PassOptionsToPackage{hyphens}{url}\usepackage{hyperref}

\usepackage{xcolor}
\usepackage{xparse}

\usepackage{array}
\usepackage{tikz}

\usepackage{breakurl}

\newcolumntype{M}[1]{>{\centering\arraybackslash}m{#1}}

\journal{International Journal of Forecasting}

\begin{document}
\begin{frontmatter}



\title{A Review of the Long Horizon Forecasting Problem in Time Series Analysis} %% Article title


\author{Hans Krupakar\corref{cor1}} %% Author name
\ead{hansk@nyu.edu}
% Author affiliation
\affiliation[snu]{organization={CSE Department, Shiv Nadar University Chennai},Department and Organization
            addressline={Rajiv Gandhi Salai (OMR)}, 
            city={Kalavakkam},
            postcode={603110}, 
            state={Tamil Nadu},
            country={India}}
\author{Kandappan V A} %% Author name
%\ead{kandappanva@snuchennai.edu.in}

\cortext[cor1]{Corresponding Author}

%% Abstract
\begin{abstract}
The long horizon forecasting (LHF) problem has come up in the time series literature for over the last 35 years or so. This review covers aspects of LHF in this period and how deep learning has incorporated variants of trend, seasonality, fourier and wavelet transforms, misspecification bias reduction and bandpass filters while contributing using convolutions, residual connections, sparsity reduction, strided convolutions, attention masks, SSMs, normalization methods, low-rank approximations and gating mechanisms. We highlight time series decomposition techniques, input data preprocessing and dataset windowing schemes that improve performance. Multi-layer perceptron models, recurrent neural network hybrids, self-attention models that improve and/or address the performances of the LHF problem are described, with an emphasis on the feature space construction. Ablation studies are conducted over the ETTm2 dataset in the multivariate and univariate high useful load (HUFL) forecasting contexts, evaluated over the last 4 months of the dataset. The heatmaps of MSE averages per time step over test set series in the horizon show that there is a steady increase in the error proportionate to its length except with \emph{xLSTM} and \emph{Triformer} models and motivate LHF as an error propagation problem. The trained models are available here: \href{https://bit.ly/LHFModelZoo}{https://bit.ly/LHFModelZoo}.
\end{abstract}

% \begin{highlights}
% \end{highlights}

%% Keywords
\end{frontmatter}



% \maketitle

\section{Introduction}
\sloppy

Time series forecasting has applications in risk management, disease propagation studies, sensor network monitoring, power plant maintenance, disaster management and prediction, construction, economics, energy and road traffic-related applications etc. (\cite{challu2023nhits, wu2021autoformer}) Most of these applications would find benefits in being able to define a horizon size that is longer and this has been shown to be true for EEG, ECG time series data, predicting clinical events, climate studies, inflation data and the financial analysis of stock markets due to the inherent nature of these datasets (\cite{thundiyiltransformer}). These benefits would come from being able to find long-term dependencies in the data and inherent co-variate analysis that isn't restricted to a short horizon. However, the presence of entangled representations increases in the context of longer horizons and this makes multi-variate correlation analysis harder than in the case of shorter horizon forecasting (\cite{wu2021autoformer}). Also, the computational complexity and memory in long-horizon forecasting models are typically dependent on the horizon size which makes longer horizon forecasting slower without GPUs (\cite{challu2023nhits}).

Statistical methods like Auto-Regressive Integrated Moving Average (\emph{ARIMA}), Exponential Smoothing (\emph{ETS}), Vector Auto-Regression (\emph{VAR}) models have been used for time series analysis for around the last 50 years but they have consistently reported a decline in performance over longer horizons (\cite{nelson1998time, cronan1975forecasting, akram2009exponential}). In the 1990's, as summarized by \cite{stock1995point}, it was widely believed that "If the variables are co-integrated, imposing this information can provide substantial improvements in forecasts over long horizons". While co-integrated systems find improvements in long horizon forecasting (LHF), most systems don't have co-integrated variables be a part of the variables considered in multi-variate time series analysis (\cite{416ffb05-26d4-3085-a9e0-04cec112cc53}). In the early 2000s, researchers in economics used long horizon forecasts to find out definitively whether one variable influences the predictions of another (\cite{clark2001evaluating}). Forecast rationality questions the functional form of the multivariate loss functions given that errors could co-occur across the variables, potentially needing larger error representations for them than the independence assumption a metric like MSE introduces (\cite{komunjer2012multivariate}). In their paper, \cite{patton2012forecast} introduce bounds to test for forecast rationality and validate them on simulated and empirical data over longer horizons, up to a horizon size of 8. Long Horizon forecasting (LHF) finds problems in computational complexity and forecast volatility in the modeling aspects of the problem (\cite{challu2023nhits}). 

In the early 21st century, forecasting methods broadly fell into Iterative Multi-Step (IMS) forecasting and Direct Multi-Step (DMS) (\cite{chevillon2007direct}) forecasting approaches wherein IMS predicted the forecast horizon one time step at a time and DMS predicted the forecast horizon altogether. Even though IMS models have a smaller variance, the error propagates through time and accumulates more and so isn't preferred for long horizon forecasting (LHF) (\cite{zeng2023transformers}). Recurrent Neural Network models overcome this error propagation using backpropagation through time (BPTT) and teacher forcing approaches. \cite{galbraith2003content} shows using simulated data that the forecast content decreases asymptotically with respect to an increase in the forecast horizon.    

 In the case of univariate time series forecasting using statistical methods, stationarity and ergodicity properties need to be satisfied (\cite{lara2021experimental}). The need for ergodicity is also visible in neural network based time series models in which the right subset of time series data taken for training gives better performance instead of the entire data. Based on findings from \cite{lara2021experimental} over tourism data, traditional models often work well in the context of shorter forecast horizons while neural network models perform well over longer horizons. This enables the use of hybrid models for the problem. Given such findings and empirical evidence, over the more recent years, time series forecasting and analysis in general in research literature has shifted from the use of traditional methods like \emph{ARIMA}, \emph{VAR}, \emph{ETS} etc. to neural forecasting approaches (and hybrids) involving multi-layer perceptrons (MLPs), recurrent neural networks (RNNs) and self-attention based approaches and the long horizon forecasting (LHF) problem has been studied extensively in the context of the use of deep learning over the past 7 years or so. Temporal CNNs are used in some works, one of which is \cite{wan2019multivariate}, where they use dilations in TCNs for multivariate forecasting and show results better than vanilla RNN LSTMs and convLSTMs. 
 
 The self-attention mechanism and transformer models for forecasting will be covered extensively in the upcoming sections of this paper. The \emph{LogTrans} model uses convolutional self-attention using causal convolutions on the key and query embeddings to leverage local context whereas traditional self-attention without local context awareness could not differentiate between anomalies, change points or parts of patterns (\cite{li2019enhancing}). Improvements to the self-attention mechanism that improve its memory and computational complexity like the use of locality sensitive hashing (LSH) in attention and causal masking to introduce sparsity show improvements in the context of extremely long sequences in language models (\cite{kitaev2020reformer, zhou2021informer}) but surprisingly doesn't as much with time series forecasting (\cite{challu2023nhits}).  Also, \cite{zhou2021informer} find similarities between the spatial aspects of the dilation operation in convolutions, de-gridding to remove dilation artifacts and length of the convolutional feature spaces (\cite{yu2017dilated}) to lengths of horizons in time series forecasting. 

Given that transformer models are able to capture long-term dependencies by construction (\cite{thundiyiltransformer}), it is important to consider the problem of long horizon forecasting (LHF) as an independent problem as well and \cite{challu2023nhits} have done this. In \cite{zeng2023transformers}, the authors challenge the efficacy of the use of transformer models for time series analysis because even though they're able to capture longer horizon dependencies, the multi-head attention mechanism in transformer models depends on permutation equivariance which could be counter-intuitive to the formulation of the time series forecasting problem as auto-regressive, although from an information point of view, it adds to the supervision provided by the data. A similar Occam's Razor (\cite{blumer1987occam}) argument about LSTMs has been made by \cite{wen2018multihorizonquantilerecurrentforecaster} when they discuss \emph{NARX RNNs} (\cite{DBLP:journals/corr/DiPietroNH17}) about how the use of a look-back input window (lagged feature inputs) along with the input at a time step helps construct skip-connections to past values of the time series data before passing them through the recurrent layer. \cite{nie2022time}, through \emph{PatchTST}'s channel independence over multi-variate series and self-supervision using patch learning, show that their transformer model performs better than MLP-based models \emph{NHITS} and \emph{D-Linear} on the same datasets (\cite{zeng2023transformers}). 

Several practitioners' tricks don't yet have ablation experiments with respect to their performances across these models and this will not be included in the review. Data augmentation techniques for time series data have been studied to try to increase the dataset sizes for forecasting that could improve the performance of long-horizon forecasting (\cite{bandara2021improving, semenoglou2023data}). Several foundation models for time series analysis with pre-trained weights have come up and all of them aren't covered in this review even though they could be applied to the forecasting problem (\cite{liang2024foundation}). Different types of evaluation metrics for time series forecasting including ones that account for seasonality and normalized scores are presented in (\cite{oreshkin2019n}). 

The following sections are organized as follows. Literature review is done based on types of models and problem formulation. The time series decomposition, windowing schemes and neural network initializations sections cover some empirical aspects of model construction. The models section discusses advances in LHF in model classes of Multi-Layer Perceptrons (MLPs), Recurrent Neural Networks (RNNs), self-attention and pyramidal self-attention. Univariate and multivariate forecasting ablations over ETTm2 minutes-based time series are covered in the experiment section. The discussion section consists of possible inferences of the experiments, motivates the LHF problem using errors per timestep plots, discusses model construction across the models and uses the performances to compare them. This review doesn't cover the theoretical guarantees and interpretability arguments used for the models and keeps the discussions to empirical analyses only.

\section{Literature Review}

Long Horizon Forecasting literature has been based on statistical models for decades, RNN-based and temporal point processes-based over the last decade and MLP and self-attention based over the last five years or so.

\textbf{Statistical Models:} \cite{vera2020long} shows an interesting result on the use of \emph{ARFIMA} over the S\&P 500 data where the realized variance was better using the \emph{ARMA} model in the shorter horizon contexts but improved quite a bit in longer horizon contexts because of \emph{ARFIMA}. In long horizon forecasting (LHF) experiments, \cite{challu2023nhits} find that the \emph{ARIMA} model (\cite{box1970distribution}) is able to perform well with some datasets and is comparable to models like \emph{NHITS} (\cite{challu2023nhits}), \emph{NBEATS} (\cite{oreshkin2019n}) and \emph{FEDformer} (\cite{zhou2022fedformer}), \emph{Autoformer} (\cite{wu2021autoformer}) but doesn't consistently do so with all the datasets.

\textbf{Recurrent Neural Network Hybrids:} The M4 competition (\cite{makridakis2018m4}) winning architecture of hybrid \emph{ETS-LSTM} uses exponential smoothing and multi-variate data across different time scales, \emph{dilated RNN LSTMs} and ensembling which worked well over a larger dataset but didn't with a smaller one (\cite{dudek2021hybrid}). While the model performed well over daily, monthly and quarterly time scales in the M4 competition, it didn't perform well over hourly and minute-based data without a sharp increase in learning rate in the latter time scales possibly due to the relative magnitude differences in time being close to zero and/or dilations (\cite{smyl2020hybrid}). A \emph{Dynamic Network Model (DNM)} is a type of belief network that takes time into account in the construction of causal graphs (\cite{dagum1992dynamic}). \cite{ismail2018improving} use LSTMs with DNMs to address long horizon forecasting (LHF) and find some correlations between their approach of using the entire dataset and Expectation Maximization and how it performs in this context. The biases are calculated using population averaging or clustering of the features relevant to \emph{DNM}. For the problem of long-term multi-variate time series forecasting, \cite{liu2020dstp} use LSTMs with a type of attention mechanism that attends to multivariate time series data spatially across time steps in two phases with and without the univariate series alongwith the exogenous variables, temporally across the hidden layer activations of the LSTM and doesn't use the spatial attention outputs in the decoder LSTM. Recently, \emph{xLSTM}s use both MLP-based and RNN-based layers with residual connections and scalar (like RNNs) and matrix memory states with covariate gradient updates per head in a multi-headed architecture. Some review papers on the use of deep learning for time series analysis outside long horizon forecasting (LHF) provide context on the use of deep learning models but don't bring up sequence length (\cite{https://doi.org/10.1049/cit2.12076, lim2021time}). There has been research in RNN architecture variants that try to address long-term dependencies and temporal aspects of the data (\cite{kag2020rnns, rusch2020coupled, li2019deep}) but they haven't yet been applied to time series forecasting.   

\textbf{Temporal Point Processes:} The \emph{Wavenet} architecture (\cite{van2016wavenet}) that connect outputs to successively iterative nodes in the past using dilated convolutions was found to provide better long-range forecasts than temporal point processes (\cite{shchur2021neural, deshpande2021long}). Dual Temporal Point Processes involve regression over time series data using two components (\cite{polanski2017forecasting}) over discretized point processes of importance. The first component involves using a recurrent neural network over the point process marks, differences in times between the mark and the previous mark and hourly index of the time series data capturing micro-scale dynamics and the second component involves using a model of counts that captures the global context. The discretization of time series data into point processes proves effective over longer horizon contexts in terms of performance but it doesn't forecast continuous time series data. (\cite{deshpande2021long})

\textbf{Multi-Layer Perceptron Models:} In Time-Series Dense Encoder \\(\emph{TiDE}) models, residual MLP layers are used throughout the architecture in an encoder decoder model. The encoder takes as input the input window, dynamic covariates after dimensionality reduction along the multivariate dimension per time step, and static attributes. The decoder is made up of multi-layer perceptron layers and a temporal layer which takes its output as input along with the embedded dynamic covariates to produce the forecasts (\cite{das2024longtermforecastingtidetimeseries}). Inspired by long-term sequence compression, given that neural network models overfit on noisy time series data, \cite{zhou2022filmfrequencyimprovedlegendre}, in their model \emph{FiLM}, use Legendre Projection, fourier transforms, boosting similar to \emph{NBEATS} (\cite{oreshkin2019n}), and low-rank matrix approximation to improve forecasting in the context of noisy test data. \emph{NBEATS} uses many MLP layers to predict both forward and backward forecasts in blocks making up stacks that follow the residual stacking principle from \emph{DenseNets} (\cite{oreshkin2019n}). \emph{NHITS} uses \emph{NBEATS}' architecture and adds sub-sampling before the MLP layers and hierarchical interpolation after and synchronizes them to improve the performance of long horizon forecasting (\cite{challu2023nhits}). \cite{zeng2023transformers} introduce \emph{D-Linear}, which uses input preprocessing similar to \emph{FEDformer}'s and \emph{Autoformer}'s to decompose it into trend and a remainder component before the MLP layers, and \emph{N-Linear}, which normalizes the input by subtracting the input by the last value of the sequence and adding it back to the output.   

\textbf{Self-Attention based Models:} \cite{chen2023long} provide a comprehensive overview of long horizon forecasting (LHF) approaches in transformer models leading up to the MLP arguments in this paper. In their work \emph{FEDformer}, \cite{zhou2022fedformer} find that a band-pass filter around fourier and wavelet transforms help encode information about long-term forecasting by improving performance in that context. The fourier transform captures frequency domain information while the use of wavelets with dimensionality reduction captures both the frequency and time domains of the signal and these act as additional data for the transformer. \emph{FEDformer} further proposes the mixture of experts’ strategies to mix the trend components extracted by moving average kernels of various kernel sizes. \emph{Autoformer} (\cite{wu2021autoformer}) uses seasonality decomposition as intermediate layers and trend cycles in a parallel connection, fourier and inverse fourier transforms instead of multi-layer perceptron layers in self-attention and time delay aggregation. Time delay aggregation adds series alignment by dividing the series into chunks and enables the auto-correlation function to focus on connections of sub-series. \cite{nie2022time} show that even though \cite{zeng2023transformers} find that \emph{FEDformer} and \emph{Autoformer} are unable to benefit from a longer input window where MLP models have, patch-based self-supervision over multi-variate data helps jointly attend to information from different patch representations of longer windows better than in the supervised learning setting. \cite{liu2024itransformerinvertedtransformerseffective} use an inverted embedding of multi-variate input time series data to use self-attention over the multi-variate dimension to extract correlations between them. 
 
\textbf{Pyramidal Self-Attention models:} \emph{Pyraformer} uses a pyramidal attention scheme and uses the hierarchies in time scales to make longer horizon forecasting easier to represent using inter-scale connections for adaptive long horizon pattern identification and intra-scale connections for relationships across time in the same scale (\cite{liu2021pyraformer}). Similar to \emph{Pyraformer}, \emph{Triformer} constructs a pyramid of triangular units and factorizes the projection matrices to use a combination of variable-agnostic self-attention layers and variable-specific weights to improve accuracy (\cite{cirstea2022triformer}).

\section{Time series decomposition}

Time series decomposition acts as feature extraction for time series data, where the data-informed sub-series enhances the performance of regression models. Statistical methods have found time series data decomposed into trend, cycle/seasonality and irregularities/residual errors using Classical, X11, SEATS (Seasonal Extraction in \emph{ARIMA} Time Series), STL (Seasonal and Trend decomposition using Loess) decomposition approaches (\cite{theodosiou2011forecasting, qi2008trend, hyndman2018forecasting}). \cite{theodosiou2011forecasting} showed the importance of decomposition techniques by using a combination of boosting and decomposition to improve the performance of the forecasting model.

\subsection{Trend}
Statistical time series models like the Box-Jenkins approach (\cite{box2015time}) depend on stationarity of the input time series data for the regression to prevent learning spurious correlations between time points based on noisy repetitions in the training data. Trend typically captures the changes in direction of the time series data, along with the corresponding magnitude of increase and decrease over time. Time series data can be either Trend-Stationary (TS), where the deterministic trend as a function of only the current time index is removed from the data or Difference-Stationary (DS), where the differences between consecutive time points are used to remove the stochastic trend from the data (\cite{qi2008trend}). \emph{NBEATS}, \emph{FEDformer} and \emph{Autoformer} use differentiable trend layers.

\subsection{Seasonality}
Classical decomposition uses averages of de-trended data across temporal gradations like months, days, hours etc. over the data to calculate seasonality but this approach fails when there is seasonality shift in the data (\cite{wen2019robuststl}). SEATS, STL and X11 decomposition techniques propose variants to the classical approach (\cite{hyndman2018forecasting}). STL and RobustSTL (\cite{wen2019robuststl}) have been shown to work better in the context of capturing longer sequences and RobustSTL's non-local seasonal filtering helps overcome seasonality shift problems prevalent in other approaches. \emph{NBEATS} and \emph{Autoformer} use differentiable seasonality layers.

\section{Windowing Schemes}

In this section, the conversion methods of a large time series dataset into input-output pairs is discussed.

\subsection{Moving Window}

Most works on time series forecasting typically use a window $I$ of t time-steps and a subsequent window $H$ of h time-steps such that $I \in \mathbb{R}^{t}$ and $H \in \mathbb{R}^h$ where $|I|=t$ and $|H|=h$. The windows are created over every temporal unit such that there is a $t-1$ times' overlap in the data for every stride of size $t$, although practitioners sometimes use different input distributions. The model takes as input $I$ and tries to forecast $H$ for every sub-series $I,H$ in the dataset, typically using gradient descent over some learnable parameters. Some models are also formulated as probabilistic inference models, in which case, the inference is made using a probability distribution.

\subsection{Forking Sequences}

In MQ-RNN, \cite{wen2018multihorizonquantilerecurrentforecaster} use a moving window scheme but for every input window $I \in \mathbb{R}^t$, create $t-1$ additional output horizons $H \in \mathbb{R}^h$. This enables lesser independence between multiple input series in the dataset within the length $t$ of every horizon and helps train models that take as input $I \in \mathbb{R}^t$ and an output horizon $\hat{H} \in \mathbb{R}^{h \times (t-1)}$. 

\section{Input Data Pre-processing}

\subsection{Normalization}

\emph{NHITS} (\cite{challu2023nhits}), \emph{TiDE} (\cite{das2024longtermforecastingtidetimeseries}), M-TCN (\cite{wan2019multivariate}), Diffusion Convolutional RNN (\cite{li2017diffusion}), \emph{Autoformer} (\cite{wu2021autoformer}), \emph{Informer} (\cite{zhou2021informer}), \emph{FEDformer} (\cite{zhou2022fedformer}), \emph{Triformer} (\cite{cirstea2022triformer}), \emph{Pyraformer} (\cite{liu2021pyraformer}) and iTransformer (\cite{liu2024itransformerinvertedtransformerseffective}) use z-score normalization ($\Tilde{y} = \frac{y-\mu}{\sigma}$ where $\mu$ is the mean and $\sigma$ is the standard deviation of $y$) of the data using the dataset's mean and standard deviation. In \emph{PatchTST} (\cite{nie2022time}), the z-score normalization is done over every input data instance and this helps mitigate the distribution shift between the train and test data. In \emph{ETS-LSTM} (\cite{smyl2020hybrid}), Smyl uses an exponential smoothing formulation and finds trend and seasonality coefficients to initialize the input for the neural network as deseasonalized and normalized. For \emph{Expectation-biased LSTMs} (\cite{ismail2018improving}), the data is differenced between the time steps, z-score normalized and imputed. In \emph{FiLM} (\cite{zhou2022filmfrequencyimprovedlegendre}), a reversible instance normalization is used which uses z-scores over the temporal dimension of the data and its inverse before the output layer. In \emph{N-Linear}, the last time step's value is subtracted from the series and added back after the forward pass.  

\subsection{Patching}

Patch-based self-supervision methods in self-attention models for time series data use masking per patch at random and this isn't suitable for time series data because it is trivial to simply interpolate between the immediate preceding and succeeding time values, making pretraining overfit. In \emph{PatchTST} (\cite{nie2022time}), the input time series data of length $L$, is divided into patches of length $P$ with a stride $S$ such that the input time series data's length goes from $L$ to approximately $\frac{L}{S}$. This helps increase the number of time steps of input and enables the use of a longer series, while reducing the computational complexity by a factor of $S$.

\subsection{Aggregation}

For \emph{NBEATS} experiments, \cite{oreshkin2019n} use aggregation of time scales of lower gradations to estimate approximate values of the higher gradation using sum or average of the series for ablation experiments in comparisons of results because of other time series models. \emph{Informer} (\cite{zhou2021informer}) uses a scalar projection of the input, local timestamp based position embeddings and global timestamp embeddings. 

%\section{Model Architectures}
\section{Models}
%In this section, different model architectures belonging to the broader classes of Multi-Layer Perceptrons (MLPs), Recurrent Neural Networks (RNNs) and Self-Attention models (Transformers) are discussed side by side. 
In this section, different model architectures are grouped by neural network model families and discussed in an ascending order of year of publication. 

\subsection{Multi-Layer Perceptron Models}

MLPs are defined by many layers, each consisting of a dot product of an input $x \in \mathbb{R}^{N \times d}$ with a learnable weight $W \in \mathbb{R}^{d \times \hat{d}}$ and an additive bias $b \in \mathbb{R}^{\hat{d}}$ followed by a point-wise non-linearity (\cite{BAUM1988193}). 

\subsubsection{NBEATS (Neural Basis Expansion Analysis for Time Series)}

Empirically, combining forecasts for the same series was found to increase robustness and improve prediction accuracy by reducing misspecification bias from lesser number of models (\cite{theodosiou2011forecasting}) and \emph{NBEATS} (\cite{oreshkin2019n}) uses this principle. Given an input window $x = [x_1, x_2,...,x_t]^T \in \mathbb{R}^t$, \emph{NBEATS} uses blocks of MLP-based layers to predict the differences between the predicted input window and the input to each block as backcast vectors and a sum of all the predicted forecast windows across stacks of blocks as the forecast vector. $v_{b,i}$ and $v_{f,i}$ are the backcast and forecast basis vectors and are defined by polynomial functions for trend, by a fourier transform for seasonality or as identity. For the first block's input $h_0=x$ and $N$ basis expansion coefficients per block, 
\begin{equation}
    \begin{split}
       h_i &= \text{relu}(W_i h_{i-1} + b_i) \quad \forall i \in \{1,2,3,4\} \\
        \hat{h}_{0,b} &= g_b(W_b h_4) \: ; \: \hat{h}_{0,f} = g_f(W_f h_4) \\
        g_b(\theta) &= \sum_{i=0}^{|\theta|} \theta_i v_{b,i} \: ; \: g_f(\theta) = \sum_{i=0}^{|\theta|} \theta_i v_{f,i} \\
        h^{(l)}_{0,b} &= h^{(l-1)}_{0,b} - \hat{h}^{(l-1)}_{0,b} \: ; \: y = \sum_l \hat{h}^{(l)}_{0,f} 
    \end{split}
\end{equation}
The trend is estimated as a polynomial of small degree $p$ where $\textbf{t}=[0, 1, 2, ..., H-2, H-1]^T/H$ and the seasonality is estimated as a discrete fourier transform.
\begin{equation}
    \begin{split}
        g^{\text{trend}}_f(\theta) &= \sum_{i=0}^p \theta \: \textbf{t}^i \: ; \:
        g^{\text{seas}}_f(\theta) = \sum_{i=0}^{\lfloor \frac{L}{2}-1\rfloor} \theta_i \cos{2\pi it} + \theta_{i+\frac{L}{2}} \sin{2\pi it}
    \end{split}
    \label{eq2}
\end{equation}
To account for the different ranges of trend and seasonality outputs in the interpretable version of the model, a sensible inductive bias is introduced into the architecture. In the ablation experiments, the identity basis function performed better.

\subsubsection{FiLM (Frequency improved Legendre Memory)}

Where the other MLP variants use the time domain directly, \emph{FiLM} (\cite{zhou2022filmfrequencyimprovedlegendre}) uses the frequency domain and a constrained memory to forecast. In the \emph{FiLM} model, an input series of length $nT$ is reversible normalized and divided into $n$ blocks that go through the Legendre Projection Unit (LPU), the fourier enhanced layer (FEL) and the LPU inverse blocks before de-normalization. The model depends on the fact that for a longer time series window, the error accumulation from history increases because of the length. The LPU contains a memory unit $C \in \mathbb{R}^N$ which accumulates the input using projection matrices $A \in \mathbb{R}^{N\times N}$ and $B \in \mathbb{R}^{N}$ where there are $N$ Legendre polynomials, as shown in equation \ref{eq3}. 
\begin{equation}
\begin{split}
    C_t &= AC_{t-1} + Bx_t \\
    A_{ij} &= (2i+1)\left\{\begin{array}{lr}
                        (-1)^{i-j}, & j \leq i\\
                        \:\:\:\:1, & j > i \\
                    \end{array}\right\} \\ 
    B_i &= (2i+1)(-1)^i
\end{split}    
\label{eq3}
\end{equation}
The fourier enhanced block (FEB) goes through a fourier transform followed by a weighted sum using the only learnable matrix $W \in \mathbb{R}^{M^{'} \times N \times N}$, which is decomposed into 3 matrices $W_1 \in \mathbb{R}^{M^{'}\times N^{'}\times N^{'}}$, $W_2 \in \mathbb{R}^{N^{'}\times N}$ and $W_3 \in \mathbb{R}^{N^{'}\times N}$ to perform low-rank matrix approximation with minor accuracy loss. The FEB includes an inverse fourier transform after the weighted transform.

\subsubsection{D-Linear and N-Linear}

\cite{zeng2023transformers} hypothesize that the improvements in performances of MLP-based models in time series forecasting is the difference between the DMS and IMS aspects of the methods. \emph{D-Linear} uses the same decomposition schemes as in \emph{FEDformer} (\cite{zhou2022fedformer}) and \emph{Autoformer} (\cite{wu2021autoformer}) and MLP layers separately on the trend and seasonality components of the input series with the final prediction being the sum of these two components. \emph{N-Linear} models simply subtract and add the last value of the series as a simple normalization technique with an MLP layer in between. These models highlight the lesser number of parameters needed for forecasting as opposed to other model families. 

\subsubsection{NHITS (Neural Hierarchical Interpolation for Time Series)}

\emph{NHITS} \cite{challu2023nhits} uses the \emph{NBEATS} architecture without the trend and seasonality based blocks and adds novel components to it to address the long horizon forecasting case specifically. Before each block performs non-linear projections onto basis functions like \emph{NBEATS} does, there is a max-pooling operation on the series input that uses different window sizes $k_b \quad \forall b \: \text{blocks}$ to enable multi-rate sampling, similar to learnable trend extraction from other models but with emphasis on the differences in the pooling window sizes. Similarly, every block uses interpolation along the temporal dimension such that for an output horizon $H_b$ for block $b$ and block forecast outputs $\theta_b$, $|\theta_b| = \lceil H_b r_b \rceil \quad \forall b \:\text{blocks}$. The interpolation operation $g(H_b,\theta_b): \mathbb{R}^{\lceil H_b r_b \rceil} \rightarrow \mathbb{R}^{H_b}$ is defined over $\tau_b = \{t+1, t+1+1/r_b, ..., t+H_b-1/r_b, t+H_b\}$ as in equation \ref{eq4}. The backcast outputs are left unchanged.   
\begin{equation}
    \begin{split}
        g(H_b, \theta_b) &= \theta_b[t_1] + (\frac{\theta_b[t_2] - \theta_b[t_1]}{t_2 - t_1})(\tau_b - t_1) \\
        t_1 &= {\arg\min}_{t \in \tau_b; t<H_b} (H_b-t) \: ; \: t_2 = t_1 + 1/r_b
    \end{split}
    \label{eq4}
\end{equation}

\subsubsection{TiDE (Time-series Dense Encoder)}

The primitives in \emph{TiDE} (\cite{das2024longtermforecastingtidetimeseries}) models are closer to computer vision models than the other MLP-based models used for forecasting. \emph{TiDE} uses a residual encoder-decoder based architecture made up of MLP units. Every input series has $s$ static exogenous variables $a \in \mathbb{R}^s$ which don't change throughout the series and $d$ dynamic exogenous variables $Y = y_{1:t} \in \mathbb{R}^{t \times d}$. The dynamic exogenous variables go through dimensionality reduction from $d$ to $\Tilde{d}$ using a residual layer. The encoder takes as input the concatenation of $x_{1:t}$, $a$ and $\Tilde{y} \in \mathbb{R}^{t \Tilde{d}}$ and uses $n_e$ residual blocks. The decoder takes as input the output of the encoder and uses $n_d$ residual blocks. The features of the dynamic exogenous variables in the output window $i_{t:t+H} \in \mathbb{R}^{H \times \Tilde{d}}$ are then stacked along-with the decoder output and put through a residual block to enable a "highway"-like connection. Finally, the input series $x_{1:t}$ is added to the output after a linear transform $\mathbb{R}^t \rightarrow \mathbb{R}^H$ in a residual skip connection and make the linear models from \cite{zeng2023transformers} subclasses of \emph{TiDE} models. 

\subsection{Recurrent Neural Networks}

RNNs are defined by a hidden layer that uses MLP output alongwith a dot product with a learnable hidden layer matrix that is learnt by backpropagation through time (BPTT). Equation \ref{eq5} shows the hidden layer formulation of RNNs through time-steps indexed by $t$ which are typically followed by an output MLP layer at each timestep.
\begin{equation}
    h_t = W_i x_t + W_h h_{t-1}
    \label{eq5}
\end{equation}
RNNs suffered from vanishing and exploding gradients problem before being able to learn sequences of certain length, of which the exploding gradients' problem was solved by gradient clipping by magnitude. The vanishing gradients' problem was addressed by gated hidden layers like LSTMs and GRUs, enabling learning longer sequences than vanilla RNNs (\cite{hochreiter1997long, chung2014empirical}). 

\subsubsection{Expectation-biased LSTMs}

In \emph{Expectation-biased LSTMs} (\cite{ismail2018improving}), two types of models show improved performances in the context of long horizon forecasting with the largest multi-variate horizon being 25 months over monthly data. It improves LHF by introducing averaging to the inputs of the RNN models. Equation \ref{eq8} specifies the \emph{Expectation-biased LSTM} models with univariate and multivariate outputs respectively, separated by $\text{OR}$. The calculation of the bias $e_{t,i}$ is done either using population averaging as specified by equation \ref{eq8} or by clustering using the silhouette method to find the number of clusters. The multivariate representation at time $t$ is then simply assigned to be the cluster centers determined by the distance function. The approach resembles the pooling-based trend extraction in MLP-based models and patching in \emph{PatchTST} except that it happens in the data space.
\begin{equation}
    \begin{split}
        \Tilde{x}_{t,i} = \left\{\begin{array}{lr}
                        x_{t,i} & t = 0\\
                        e_{t,i} & t > 0\\
                    \end{array}\right\} \: &\text{OR} \: [x_{t,i}, e_{t,i}] \\
        \hat{y}_{t+1} = \text{LSTM}(\Tilde{x}_{t,1}, \Tilde{x}_{t,2},...,\Tilde{x}_{t,n}, \hat{y}_t) \: &\text{OR} \: f(\text{LSTM}(\Tilde{x}_{t,1}, \Tilde{x}_{t,2},...,\Tilde{x}_{t,n})) \\
        e_{t,i} = \beta(t)x_{t,i} + (1-\beta(t))\mu_{t,i} \:&;\: \mu_{t,i} = \frac{1}{N} \sum_{j=0}^N x_{t,j} \\
    \end{split}
    \label{eq8}
\end{equation}

\subsubsection{DSTP-RNN}

In \emph{DSTP-RNN} (\cite{liu2020dstp}), given the time series data input $x \in \mathbb{R}^{T}$ and corresponding exogenous variables $Y \in \mathbb{R}^{T\times N}$ of $N$ variables per timestep, $Y$ goes through a one-layer LSTM encoder with spatial and temporal attention modules and a similar decoder. The first spatial attention layer uses a softmax over $\mathbb{R}^{N}$ and is applied to the input of the next spatial attention layer. The second layer takes as input $[\Tilde{y},x] \in \mathbb{R}^{T\times (N+1)}$ and applies the same attention-based operation as the first layer on the inputs.
\begin{equation}
    \begin{split}
        h^{(1)}_{t} &= f_e(h^{(1)}_{t-1}, Y_t)\\
        \alpha_t &= \text{softmax}(v_i \tanh{(W_i[c^{(1)}_{t-1},h^{(1)}_{t-1}]+U_iz_{i,t}+b_i)}) \\
        \Tilde{y}_t &= [\alpha_t^1z_{i,t}^1,\alpha_t^2z_{i,t}^2,...,\alpha_t^Nz_{i,t}^N] \:; \: z_{i,t} = \left\{\begin{array}{lr}
              Y_t & i=0\\
              {[}x_t,\Tilde{y}_t{]} & i=1
        \end{array}\right\} 
    \end{split}
\end{equation}
where $v_i, b_i \in \mathbb{R}^{T}, W_i \in \mathbb{R}^{T\times 2m}, U_i \in \mathbb{R}^{T\times T}$. While the first two spatial attention layers attend to distributions over the inputs, the third attention layer attends to distributions over the hidden layer activations $h^{(1)}_t \in \mathbb{R}^{p}$ for every timestep $t$ of the RNNs.
\begin{equation}
    \begin{split}
        \beta_t = \text{softmax}(&v_2 \tanh{(W_2[c^{(2)}_{t-1},h^{(2)}_{t-1}]+U_2h^{(1)}_t+b_2)}) \\
        &\Tilde{h}_t = [\beta_t^1h_t^1,\beta_t^2h_t^2,...,\beta_t^Th_t^T] \\
        &c_t = \sum_{j=0}^T \beta_t^j h_j^{(1)}
    \end{split}
\end{equation}
where $W_2 \in \mathbb{R}^{q\times 2p}, U_2 \in \mathbb{R}^{q\times q}, h^{(1)}_t \in \mathbb{R}^{q\times q}, v_2, b_2 \in \mathbb{R}^q$. After the encoder LSTM and attention layers, the decoder takes the attention context vector $c$ and input series $x$ as input and uses a single RNN layer to produce outputs. The outputs $d \in \mathbb{R}^{T \times p}$ are concatenated with the attended encoder vector $c \in \mathbb{T\times q}$ and an affine transform of the size of the output horizon is used.
\begin{equation}
    \begin{split}
        d_t &= f_d(d_{t-1}, \Tilde{w}^T[x_t, c_t]+\Tilde{b}) \\
        \hat{y}_{T+1},\hat{y}_{T+2},...\hat{y}_{T+\tau} &= v_o^T(W_y[d_t,c_t]+b_y) + b^{'}_y
    \end{split}
\end{equation}
where $\Tilde{w} \in \mathbb{R}^{q+1}, \Tilde{b} \in \mathbb{R}, W_y \in \mathbb{R}^{p\times(p+q)}, b_y \in \mathbb{R}^p, v_o \in \mathbb{R}^{\tau \times p}, b^{'}_y \in \mathbb{R}^\tau$. The spatial attention layers address multivariate forecasting more than other models do in this review by model formulation. 

\subsubsection{xLSTM}

\emph{xLSTM} consists of two types of residual layers, sLSTM and mLSTM consisting of scalar and matrix memory respectively, with a covariance update rule for the matrix memory whose weight uses a third dimension for per-head covariance scores. The input and forget gates of the LSTM use expontential gating and stabilization which makes those gate outputs positive. The sLSTM layer uses LSTM's memory mixing and group normalization for each head. For input, forget, cell and output gates of LSTMs $i_t$, $f_t$, $c_t$ and $o_t$, stabilization state $m_t$, sLSTM gates are defined with a superscript s and for non-recurrent input, forget and output gates $\Tilde{i_t}=w_i^Tx_t + b_i, \Tilde{f_t}=w_f^Tx_t + b_f, \Tilde{o_t}=W_ox_t + b_o$, mLSTM layers are defined with a superscript m.
\begin{equation}
    \centering
    \begin{split}
        i^s_t &= e^{i_t} \\
        f^s_t &= \sigma(f_t) \text{ OR } e^{f_t} \\ 
        o^s_t &= \sigma(o_t) \\
        m_t &= \max(\log(f^s_t)+m_{t-1}, \log(i^s_t))\\
        i^s_t &= e^{i_t-m_t} \\
        f^s_t &= e^{\log(f^s_t) - (m_t-m_{t-1})}\\ \\
        c^s_t &= f^s_t c^s_{t-1} + i^s_t \varphi(c_t) \\
        n^s_t &=  f^s_t n^s_{t-1} + i^s_t \\
        h^s_t &= \sigma(o_t) \frac{c^s_t}{n^s_t}
    \end{split} \hspace{0.5cm} \Bigg| \hspace{0.5cm}
    \begin{split}
        i^m_t &= e^{\Tilde{i_t}} \\
        f^m_t &= \sigma(\Tilde{f_t}) \text{ OR } e^{\Tilde{f_t}} \\ 
        o^m_t &= \sigma(\Tilde{o_t}) \\
        q_t &= W_q x_t + b_q \\
        k_t &= \frac{1}{\sqrt{d}} W_k x_t + b_k \\
        v_t &= W_v x_t + b_v \\
        n^m_t &= f^m_t n^m_{t-1} + i^m_t k_t \\
        C^m_t &= f^m_t C^m_{t-1} + i^m_t v_t k_t^T \\
        h^m_t &= o^m_t \odot \frac{C^m_t q_t}{\max(|{n^m_t\:}^T q_t|, 1)} 
    \end{split}
\end{equation}
The layers can use causal convolution before the LSTM calculations and for this review, a strided version of this has been implemented and shows improvements in the performances of LHF in the multivariate forecasting context.  

\subsection{Self-Attention Models}

The input $\hat{X} = [\hat{x}_1,\hat{x}_2,...,\hat{x}_t]^T \in \mathbb{R}^{t \times (d+1)}$ with $d$ exogenous variables goes through multi-head self-attention sub-layers, each of which transforms the input into key $K \in \mathbb{R}^{t \times k}$, query $Q \in \mathbb{R}^{t \times k}$ and value $V \in \mathbb{R}^{t \times v}$ matrices. The self-attention matrix $A_h \in \mathbb{R}^{t \times t}$ is obtained by a dot product over the dimension of length $k$ such that the dimensionality of the result covers correlations over the input dimension of length $t$. $A_h$ is then masked by an identity-based lower triangular mask $M$ with $-\infty$ in the upper triangle of the matrix using a Hadamard product. The resultant $[O_h \quad \forall h \in \{1,2,3,...,h\}]$ matrices over $h$ transformer heads are concatenated and linearly projected by $W_O \in \mathbb{R}^{hv\times t}$ and ReLU activation is applied (\cite{vaswani2017attention}). 
\begin{equation}
\begin{split}
    A_h &= \text{softmax}(\frac{Q_h K_h^T}{\sqrt{k}}) \\
    O_h &= (A_h \odot M) V_h \quad \forall h \in [1,H]
\end{split} 
\end{equation}

\subsubsection{LogSparse Transformer}

In \emph{LogSparse transformers} (\cite{li2019enhancing}), if the input time series window is represented by $I=\{x_i \quad \forall i \in \{1,2,3,...,t\}\}$ of length t, every time point $x_i$ is concatenated with temporal co-variates $z_{i+1} \in \mathbb{R}^d$ that are known for the entire period t such that the input $\hat{x}_i=\{(x_{i-1}, z_{i}) \quad \forall i \in \{1,2,3,...,t\}\}$ uses temporal co-variates from the time-step of forecast as input. To help with dependencies in time of the time-steps after events, holidays, trends etc., causal convolutions for self-attention is introduced. Motivated by random sparsity patterns in attention maps, the self-attention is further restricted from all past time-steps to all past $\log L$ steps, reducing the computational complexity from $O(L^3)$ to $O(L(\log L)^2)$ with some minor assumptions around the dimensionality differences.

\subsubsection{Informer}

Based on findings around a small percentage of self-attention tokens contributing more to the attention and a long tail distribution, \cite{zhou2021informer} use KL-divergence to measure the deviation $M$ per query given the key $K$ from a trivial case where all self-attention scores are uniformly distributed. Here, the top-n queries $\Tilde{Q}$ are chosen based on $M$ because a higher value of $M$ indicates a larger deviation from the uniform distribution case and lower sparsity. 
\begin{equation}
    \begin{split}
        M(q_i, K) &= \ln{\sum_{i=0}^{k} e^{\frac{q_i k_j^T}{\sqrt{k}}}} - \frac{1}{k} \sum_{i=0}^{k} q_i k_j^T \\
        A_h &= \text{softmax}(\frac{\Tilde{Q}_h K_h^T}{\sqrt{k}}) \\
    \end{split}
\end{equation}

Further, self-attention distilling is introduced which 1) uses a $1\times1$ convolution, ELU and max-pooling with stride 2 to halve the dimensionality and 2) drops one layer at a time to reduce the dimensionality. The encoder takes an embedding of a long series as input and the decoder takes smaller series as input compared to the encoder, padded with zeroes as placeholders for the forecasts. Where \emph{LogSparse Transformer} depends on learning-based sampling of weights as a function of length, \emph{Informer} uses KL-divergence and directly addresses sparsity as a function of magnitude.

\subsubsection{Autoformer}

In \emph{Autoformer}s (\cite{wu2021autoformer}), the input time series window of length $t$ goes through differentiable trend $x_T = [\text{avgpool}(\text{padding}(x_{\frac{t}{2}:t})), \text{mean}(x_{1:t})] \in \mathbb{R}^{t}$ where the mean makes sure the trend isn't imposed over the last few time-steps and seasonality $x_S = [x_{\frac{t}{2}:t} - \text{avgpool}(\text{padding}(x_{\frac{t}{2}:t})), 0] \in \mathbb{R}^{t}$ where the zeroes make sure of something similar with seasonality. The self-attention mechanism in transformer architectures is replaced by an autocorrelation operation inspired by stochastic process theory. The autocorrelation of time series data $x_{1:t}$ at $\tau$ is estimated as confidence scores between $x_t$ and $x_{t-\tau}$ and over all the time-steps based on FFT's and IFFT's logarithmic complexity and choosing the top activations based on a logarithmic number of time points. The use of time-delay aggregation in the calculation of autocorrelation makes the network focus on time blocks and trends between them instead of just point-wise self-attention, using a roll operation that rotates the series clockwise.

\subsubsection{FEDformer}

To make sure long sequences are modelled well, \cite{zhou2022fedformer} both ignore high frequency components of fourier transforms of key $K$, query $Q$ and value $V$ and retain them to avoid noise overfitting and to represent important events respectively, by selecting a constant number of modes $M$. \emph{FEDformer} uses residual connections in all the attention layers, DFT and fixed Legendre wavelet basis transformations (in separate models) to augment the self-attention process. To provide expressivity to moving window based trend extraction, they use a mixture of experts of trend components by applying a softmax over pooling functions. Together, every encoder is made up of 1) frequency-enhanced block and 2) multi-layer perceptron layer with the decoder adding a frequency-enhanced attention block in between the two layers of the encoder. Every fourier-based block first goes through a Hadamard product of the randomly selected $M$ queries $\Tilde{Q}$ with a parameterized kernel $R$ initialized randomly whereas every wavelet-based block uses Legendre wavelets basis decomposition and splits the results into high frequency, low frequency and the remaining parts and uses three fourier-based blocks on them.   
\begin{equation}
    \begin{split}
        \Tilde{Z} &= \text{Select}(\digamma(Z)) \qquad Z \in \{K,Q,R\} ; \digamma \in \{\text{fourier}, \text{wavelet}\} \\
    A_h &= \sigma(\Tilde{Q_h} \Tilde{K}_h^T) \\
    O_h &= \digamma^{-1}(\text{padding}(A_h V_h)) \quad \forall h \in [1,H]
        \end{split}
        \label{eq}
\end{equation}

The wavelet attention block differs from the fourier attention block in that the self-attention mechanism is replaced by the same operations as the wavelet-based block, substituting the three fourier-based blocks with three fourier attention blocks as described by equation \ref{eq}. All the fourier and wavelet transforms are in the complex domain. \emph{FEDformer} uses a set of pre-selected fourier bases during inference for quicker forecasting. The training time of \emph{FEDformer} was the longest out of all the MLP-based and self-attention models in this review.

\subsubsection{PatchTST}

\emph{PatchTST} (\cite{nie2022time}) uses an additive position encoding scheme over the latent space of the input using an affine transform and this is the input of the encoder-only transformer model which has batchnorm and MLP layers. For the self-supervised pretraining, some patches are selected uniformly at random and are filled with zero values to train the model to reconstruct them. The architecture hasn't been modified from the baseline self-attention model and the patches are split uniformly across the attention heads in the architecture. The self-supervised training helps improve the performance of forecasts of longer horizons proportionate to the length of the horizon.

\subsection{Pyramidal Self-Attention Models}

Pyramidal self-attention models exploit the hierarchical nature of time series gradations and use inter-scale connections as an N-ary tree to improve long horizon forecasting performance using coarser and finer feature representations.

\subsubsection{Pyraformer}

For temporal gradations $s=1,2,...,S$ in increasing order of hierarchy with $C^{(s)}$ units per parent node, the number of self-attention edges $\mathbb{N}_{n}^{(s)}$ of a node $n_j^{(s) } \quad \forall j \in \{1,2,3,...L^{(s)}\}$ of $L^{(s)}$ nodes of scale $s$ is given by the union of the set of adjacent nodes $\mathbb{A}_n^{(s)}$, the set of children nodes $\mathbb{C}_n^{(s)}$ and the set of parent nodes $\mathbb{P}_n^{(s)}$. For simplicity of implementation, $C^{(s)} = C^{(t)} \quad \forall t \in \{1,2,3,...S\}$. If every node $n_j$ has $A_j$ adjacent vertices where $|A_j|=L-j$,    
\begin{equation}
\begin{split}
    \mathbb{N}_{n}^{(s)} &= \bigcup \left\{\begin{array}{l}
                        \mathbb{A}_n^{(s)} =  \{n_j^{(s)}: |j-n|<\frac{A_j-1}{2}, 1 \leq j \leq \frac{L^{(s)}}{C^{(s-1)}}\},\\
                        \mathbb{C}_n^{(s)} =  \{n_j^{(s-1)}: (n-1)C^{(s)} \leq j \leq nC^{(s)}, s\neq 1\},\\
                        \mathbb{P}_n^{(s)} =  \{n_j^{(s+1)}: j = \lceil \frac{L^{(s)}}{C^{(s)}}\rceil, s\neq S\}
                    \end{array}\right\} \\ 
    A_h &= \text{softmax}_{l \in \mathbb{N}_{n}^{(s)}}(\frac{Q_l K_h^T}{\sqrt{k}}) \\
    O_h &= (A_h \odot M) V_l \quad \forall h \in [1,H], l \in \mathbb{N}_{n}^{(s)}
\end{split} 
\label{eq6}
\end{equation}

The pyramidal attention is implemented as $S$ convolutional layers of kernel size and stride $C$ in a bottleneck block that reduces the dimensionality. The two decoder self-attention layers both take the encoder output and concatenate it with the layer outputs to extract features well from the pyramidal self-attention (\cite{liu2021pyraformer}). 

\subsubsection{Triformer}

In \emph{Triformer} (\cite{cirstea2022triformer}), the input series $x_L$ of length $L$ is divided into $S$ patches such that every patch $x_p=\{x_{(p-1)S + 1}, x_{(p-1)S + 2},..., x_{pS}\}$ of length $|x_p| = \frac{L}{S}$ and alongwith its parent forms a single triangle in the architecture. The self-attention mechanism is defined in equation \ref{eq7}. To enable connections between the parent nodes of parents (called pseudo timestamps), a recurrent connection with learnable parameters $\Theta_1, \Theta_2, b_1, b_2$ is used.
\begin{equation}
    \begin{split}
    \mathbf{T}_p &= (\text{softmax}(\frac{\mathbf{T}_p^{(i)} (x_p^{(i)} K^T)}{\sqrt{k}})  \odot M) (x_p V) \quad \forall i \in \{1,2,...,S\} \\
    \mathbf{T}_{p+1} &= \tanh{(\Theta_1 \mathbf{T}_p + b_1)} + \text{sigmoid}(\Theta_2 \mathbf{T}_p + b_2) + \mathbf{T}_{p+1}
    \end{split}
    \label{eq7}
\end{equation}
The size of every $(l+1)^{th}$ layer with respect to every $l^{th}$ layer is reduced by a factor of $\frac{1}{S}$. For every layer, a neural network's output over all the pseudo timestamps $T_p \quad \forall p \in \{1,2,...,\frac{L}{S}\}$ is taken as the input to the next layer. To enable variable-specific \emph{Triformer} blocks, the key and value projection matrices $P=K,V \in \mathbb{R}^{t \times d}$ are each factorized into three matrices $L_P \in \mathbb{R}^{t \times a}, B_P \in \mathbb{R}^{a \times a}, R_P \in \mathbb{R}^{a \times d}$ where the left and right matrices $L_P, R_P$ are kept fixed across all the layers but $B_P$ of smaller dimension is learned per-patch. Intuitively, where \emph{Pyraformer} describes its architecture using a top-down strategy, \emph{Triformer} constructs the hierarchy using graph primitives bottom-up.  

\section{Experiment}
This section consists of the dataset description, dataset splits and differences between the cited work, multivariate forecasting variants and ablation experiments over the models.
\subsection{ETTm2 Dataset}
The ETTm2 dataset consists of Electricity Transformer Temperature data in 15-min intervals in one of the two regions of the dataset in China, collected by the authors of \emph{Informer} \cite{zhou2021informer}. The dataset consists of high, medium and low variants of useful and useless loads of electricity and oil temperature of the transformer. The High UseFul Load (HUFL) was chosen for univariate forecasting and all of them for the multivariate forecasting ablation experiments.
\subsection{Dataset Splits}
The first 16 months' data is used as the training set, the next 4 months' data as the validation set and the last 4 months' data as the test set, as opposed to the 12/4/4 months' convention from Zhou et al. In table \ref{table1}, the split column 7:1:2 represents the use of 14 months' data for training as opposed to the 16 months' data used for the results. The changes in performance reported in the ablation experiments can be attributed to this difference in dataset splits. The horizon sizes $H \in \{96,192,336,720\}$ taken for the ablation experiments, corresponding to 2, 4, 7 and 15 days respectively, following convention from the cited work.
\subsection{Multivariate Forecasting Variants}
\emph{NBEATS} and \emph{NHITS} models are trained using a one-dimensional input and one-dimensional output where the dataset variables are each part of the mini-batch whereas the other models in this study take as input all the variables in the dataset at once.   
\subsection{Ablations}
The results tabulated in Table \ref{table1} for multivariate forecasting and Table \ref{table2} for univariate forecasting have been configured using hyper-parameter configurations that worked well with respect to the Mean Average Error (MAE). The least error scores are highlighted in bold, and the second least scores are underlined. Some models preferred a larger input size corresponding to the horizon size whereas some like \emph{TiDE} performed better when the input size stayed lower than the horizon sizes as they increased, as is highlighted by the model zoo. The training times of \emph{PatchTST}, \emph{FEDformer} and \emph{Autoformer} were the longest in descending order. \emph{xLSTM} was implemented with strided convolution of length equal to the kernel size and repeated features to maintain the length for the multivariate forecasting task. \emph{TiDE} models weren't implemented using time units and distance to holidays as static variables as mentioned in the paper. The \emph{Pyraformer} model uses deep strided convolutions imitating time scales in the feature space, and pyramidal self-attention, which is why it performs slightly better than \emph{Informer} in the multivariate case for a horizon of size 96 and in the univariate ablations. \emph{Triformer} models were more than 30 times as big as the second largest models, \emph{FEDformer}. The pre-training in \emph{PatchTST} didn't include other time series datasets. Most of the models found improvements because of early stopping, indicating the lack of supervision because of the time series data. The code for the pretrained models is available here: \href{https://github.com/hansk0812/Forecasting-Models}{https://github.com/hansk0812/Forecasting-Models} for all the models except \emph{PatchTST}. 
\begin{table}
    \begin{adjustbox}{width=\columnwidth,center}
    \begin{tabular}{p{2.5cm}ccccccccccc} \toprule
         \multirow{2}{2.5em}{\centering Family} & \multirow{2}{2.5em}{\centering Model} & \multicolumn{2}{c}{H=96}  & \multicolumn{2}{c}{H=192} & \multicolumn{2}{c}{H=336} & \multicolumn{2}{c}{H=720} & \multirow{2}{1.5em}{\centering 7:1:2} & \multirow{2}{2.5em}{\centering $\Delta$} \\
         \cmidrule(lr){3-4}\cmidrule(lr){5-6}\cmidrule(lr){7-8}\cmidrule(lr){9-10}
         & & MSE & MAE & MSE & MAE & MSE & MAE & MSE & MAE & & \\
         \midrule \\
            \multirow{5}{2.5cm}{Multi-Layer Perceptrons} & NBEATS & 0.12607 & 0.25389 & 0.15852 & 0.29161 & 0.20527 & 0.33431 & 0.21008 & 0.32999 & $\checkmark$ & $+25.12\%$ \\
            & NHITS & 0.12632 & 0.25725 & 0.15684 & 0.28402 & \underline{0.18263} & \underline{0.30519} & \textbf{0.20643} & \textbf{0.32390} & $\checkmark$ & $+23.68\%$ \\
            & TiDE & 0.16967 & 0.24948 & 0.22099 & 0.28789 & 0.26809 & 0.32578 & 0.32626 & 0.44437 & $\checkmark$ & $-2\%$ \\
            & N-Linear & \underline{0.11258} & \underline{0.24178} & \textbf{0.13692} & \textbf{0.26834} & \textbf{0.16759} & \textbf{0.30049} & \underline{0.21101} & \underline{0.34598} & $\times$ & $+7.9\%$\\
            & D-Linear & 0.19071 & 0.26388 & 0.23998 & 0.29683 & 0.28435 & 0.32796 & 0.35558 & 0.37678 & $\times$ & $+0.4\%$\\
             & FiLM & 0.18783 & 0.26973 & 0.248902 & 0.30799 & 0.28109 & 0.33304 & 0.35619 & 0.38421 & $\checkmark$ & $-3.95\%$ \\
            & & & & & & & & \\
            RNNs & xLSTM & 0.15198 & 0.28063 & 0.21677 & 0.33933 & 0.32690 & 0.43300 & 0.31646 & 0.43124 & $\Ydown$ & $\Ydown$ \\
            & & & & & & & & \\
            \multirow{2}{2.5cm}{Self-Attention} & Informer & 0.15096 & 0.28938 & 0.17421 & 0.30644 & 0.26840 & 0.39117 & 0.35977 & 0.44505 & $\times$ & $+62.64\%$ \\
            & Autoformer & 0.13274 & 0.26023 & 0.15552 & 0.27957 & 0.20212 & 0.32277 & 0.22771 & 0.34023 & $\times$ & $+31.5\%$ \\
            & FEDformer & 0.12952 & 0.25209 & \underline{0.14705} & \underline{0.27064} & 0.19777 & 0.31976 & \underline{0.21476} & \underline{0.34550} & $\checkmark$ & $+28.58\%$ \\
            & PatchTST & 0.17009 & 0.26088 & 0.22571 & 0.29571 & 0.28316 & 0.32948 & 0.35941 & 0.38007 & $\times$ & $-0.87\%$ \\ 
            & & & & & & & & \\
            \multirow{2}{3cm}{Pyramidal Self-Attention} & Pyraformer & 0.14285 & 0.28155 & 0.18989 & 0.33351 & 0.29432 & 0.40333 & 0.49207 & 0.53307 & $\times$ & $+64.18\%$ \\
            & Triformer & \textbf{0.11745} & \textbf{0.23287} & 0.18107 & 0.30210 & 0.23462 & 0.34103 & 0.25790 & 0.36652 & $\Ydown$ & $\Ydown$ \\
         & & & & & & & & \\
         \bottomrule 
    \end{tabular}
    \end{adjustbox}
        \caption{ETTm2 multivariate forecasting metrics for different horizons H=\{96,192,336,720\}: a $\checkmark$ indicates that the cited work uses a 7:1:2 train-validation-test set splits as opposed to the 6:2:2 used for these experiments. $\Delta$ indicates the deviation between the scores from the cited work and the ablation experiments results in percentage. In the $7:1:2$ column, a $\times$ indicates that the deviation is reported on 6:2:2 train-val-test splits whereas a $\checkmark$ indicates that the cited work used a 10\% larger training dataset. A $\Ydown$ in the $\Delta$ column indicates that this paper is the first one that reports these scores for ETTm2, to our knowledge. The lowest error scores are highlighted in bold and the second lowest scores are highlighted using an underline.} 
    \label{table1}
\end{table}
\begin{table}
    \begin{adjustbox}{width=\columnwidth,center}
    \begin{tabular}{p{2.5cm}ccccccccc}\toprule
         \multirow{2}{2.5em}{\centering Family} & \multirow{2}{2.5em}{\centering Model} & \multicolumn{2}{c}{H=96}  & \multicolumn{2}{c}{H=192} & \multicolumn{2}{c}{H=336} & \multicolumn{2}{c}{H=720} \\
         \cmidrule(lr){3-4}\cmidrule(lr){5-6}\cmidrule(lr){7-8}\cmidrule(lr){9-10} 
         & & MSE & MAE & MSE & MAE & MSE & MAE & MSE & MAE \\
         \midrule \\
         \multirow{6}{2.5cm}{Multi-Layer Perceptrons} & NBEATS & 0.16835 & 0.30927 & 0.21007 & 0.35075 & 0.27021 & 0.38455 & 0.32533 & 0.43065 \\
         & NHITS & 0.16725 & 0.30139 & 0.20292 & 0.34051 & \underline{0.22374} & \underline{0.35326} & 0.29218 & 0.41082 \\
         & TiDE & 0.15959 & 0.29649 & 0.21847 & 0.35038 & 0.28029 & 0.32501 & 0.35793 & 0.37910 \\ 
         %0.27002 & 0.35289 & 0.27364 & 0.39642 & 0.31321 & 0.42071 & 0.32565 & 0.42388 \\
         & D-Linear & 0.18155 & 0.31583 & 0.23618 & 0.36144 & 0.29184 & 0.40321 & 0.33263 & 0.42905 \\
         & N-Linear & 0.17488 & 0.31479 & 0.21110 & 0.34651 & 0.24839 & 0.37705 & \underline{0.28170} & \underline{0.40765} \\
         & FiLM & 0.16905 & 0.30755 & \textbf{0.19189} & \textbf{0.32809} & \textbf{0.22749} & \textbf{0.36006} & \textbf{0.26652} & \textbf{0.39694} \\
         & & & & & & & & \\
         RNNs & xLSTM & 0.18287 & 0.32011 & 0.22599 & 0.35571 & 0.31046 & 0.41855 & 0.32941 & 0.42872 \\   
         & & & & & & & & \\ 
         \multirow{4}{2.5cm}{Self-Attention} & Informer & 0.18885 & 0.26549 & 0.26187 & 0.31705 & 0.33662 & 0.36783 & 0.42206 & 0.42381 \\
         & Autoformer & \underline{0.15051} & \underline{0.28341} & 0.20248 & 0.33812 & 0.27201 & 0.40082 & 0.33310 & 0.43064 \\
         & FEDformer & \textbf{0.11973} & \textbf{0.24940} & 0.21771 & 0.35352 & 0.32424 & 0.36163 & 0.33840 & 0.44271 \\
         & PatchTST & 0.31986 & 0.39899 & 0.41276 & 0.44777 & 0.44266 & 0.46468 & 0.57651 & 0.54554 \\
         & & & & & & & & \\ 
         \multirow{2}{3cm}{Pyramidal Self-Attention} & Pyraformer & 0.16259 & 0.30421 & \underline{0.19605} & \underline{0.33487} & 0.23098 & 0.37042 & 0.35839 & 0.46661 \\ 
         & Triformer & 0.20780 & 0.33864 & 0.26466 & 0.37744 & 0.33412 & 0.42149 & 0.36018 & 0.44304 \\
         & & & & & & & & \\
\bottomrule
    \end{tabular}
    \end{adjustbox}
        \caption{ETTm2 univariate HUFL forecasting metrics for different horizons H. The lowest error scores are highlighted in bold and the second lowest scores are highlighted using an underline.} 
    \label{table2}
\end{table}
\begin{table}
    \centering
    \begin{adjustbox}{width=\columnwidth,center}
    \begin{tabular}{ccccccccc}\toprule
        \multirow{3}{2.5em}{\textbf{Series}} & \multicolumn{4}{c}{\textbf{NHITS}} & \multicolumn{4}{c}{\textbf{N-Linear}} \\ 
        & \multicolumn{2}{c}{\textbf{Multivariate}} & \multicolumn{2}{c}{\textbf{Univariate}} & \multicolumn{2}{c}{\textbf{Multivariate}} & \multicolumn{2}{c}{\textbf{Univariate}} \\
        \cmidrule(lr){2-3}\cmidrule(lr){4-5}\cmidrule(lr){6-7}\cmidrule(lr){8-9}
        & MSE & MAE & MSE & MAE & MSE & MAE & MSE & MAE \\
        \midrule
        HUFL & 0.30724 & 0.42264 & 0.31641 & 0.42075 & 0.30400 & 0.42012 & 0.28170 & 0.40765 \\
        HULL & 0.30072 & 0.40812 & 0.28516 & 0.39358 & 0.25274 & 0.38149 & 0.27103 & 0.39843 \\
        MUFL & 0.12878 & 0.27227 & 0.11827 & 0.25779 & 0.24276 & 0.39884 & 0.11666 & 0.26260 \\
        MULL & 0.29476 & 0.41883 & 0.29674 & 0.42063 & 0.25144 & 0.38853 & 0.25943 & 0.39644 \\
        LUFL & 0.06692 & 0.20535 & 0.07193 & 0.21379 & 0.08148 & 0.23055 & 0.04442 & 0.16646 \\
        LULL & 0.01147 & 0.08417 & 0.01047 & 0.08003 & 0.04923 & 0.17170 & 0.00840 & 0.07204 \\
        OT & 0.31456 & 0.44267 & 0.28888 & 0.42905 & 0.30724 & 0.41755 & 0.31811 & 0.44239 \\
        \midrule
        \textbf{Average} & \textbf{0.20643} & \textbf{0.32390} & \textbf{0.19826} & \textbf{0.31651} & \textbf{0.21094} & \textbf{0.34597} & \textbf{0.18568} & \textbf{0.30657} \\
        \bottomrule
    \end{tabular}
    \end{adjustbox}
    \caption{Over the \emph{NHITS} and \emph{N-Linear} models, per-series metrics MSE and MAE of ETTm2 series are calculated and averaged for predictions of horizon size 720 over univariate and multivariate models for empirical analysis of single-input and multi-input multivariate MLP models}
    \label{table3}
\end{table}
\section{Discussion}
\subsection{Empirical Observations}
The models \emph{Informer}, \emph{Autoformer} and \emph{FEDformer} forecast a part of the input similar to an Autoencoder and find performances that are better than some models that don't. Given that the representative capabilities of neural network models depend on inputs of higher dimensionalities, corresponding feature dimensions and large datasets, objective comparisons of feature spaces seem difficult with just ETTm2 and other similarly small time series forecasting datasets. Table \ref{table3} highlights the differences in performance between MLP-based univariate and multivariate forecasting model variants where \emph{NHITS}, different only because of amounts of data, shows that the models with the same hyperparameter configurations are able to learn representations similar in terms of the losses, highlighting the importance of feature space construction over the number of parameters. \emph{N-Linear}, on the other hand, shows overall improvement in performance because of the univariate formulation of the problem, highlighting that for a dataset the size of ETTm2, the simpler the model, the lesser they could benefit from a multivariate problem formulation. Table \ref{table4} shows the per-timestep predicted horizon errors of different types of models in the case where the horizon size is 720. The associated heatmap that is more yellow where the errors are higher and it is clear that most all the models except \emph{Triformer} and \emph{xLSTM} struggle because of the length of the horizon. The fact that \emph{Triformer} is able to overcome this limitation is because of the hierarchical construction of the feature space. Both \emph{Triformer} and \emph{xLSTM} don't perform as well as the better models do but the sLSTM and mLSTM layers in \emph{xLSTM} seem to reverse the dependence of performance on length when the models find the best hyperparameters over the ETTm2 test set. \emph{xLSTM}'s strided convolution-based implementation is visible in its multivariate forecasting heatmap in table \ref{table4}, corresponding to the stride value. Interestingly, \emph{NHITS} also shows a similar pattern in the multivariate error heatmap even though the pooling and interpolation operations happen in the feature space. The importance of pretraining in \emph{PatchTST} is highlighted by the differences in performance between the univariate and multivariate models and using the entire dataset to pretrain in the univariate context could improve the performance in that context. \emph{Expectation-biased LSTMs} and \emph{DSTP-RNN} models haven't been tested against the standard long horizon forecasting datasets or the appropriate window sizes and their improvements haven't been compared with the others' directly in this review.
\begin{table}[t]
    \centering
    \resizebox{\textwidth}{!}{
    \begin{tabular}{ccM{5cm}cM{5cm}M{0.4cm}}
        \toprule
        \multirow{2}{2.5em}{\textbf{Model}} & \multicolumn{2}{c}{\textbf{Univariate}} \hspace{2cm} & \multicolumn{2}{c}{\textbf{Multivariate}} \hspace{2cm} & \\ 
        & \textbf{MSE} & \textbf{Heatmap} & \textbf{MSE} & \textbf{Heatmap} & \\
        \midrule
        Pyraformer & 0.35839 & \begin{tikzpicture} \node[inner sep=0pt] {
            \includegraphics[width=5cm,height=1cm]{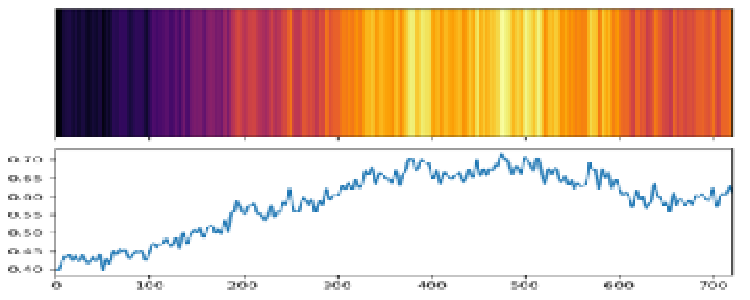}
        };\end{tikzpicture} & 0.49207 & \begin{tikzpicture} \node[inner sep=0pt] {
            \includegraphics[width=5cm,height=1cm]{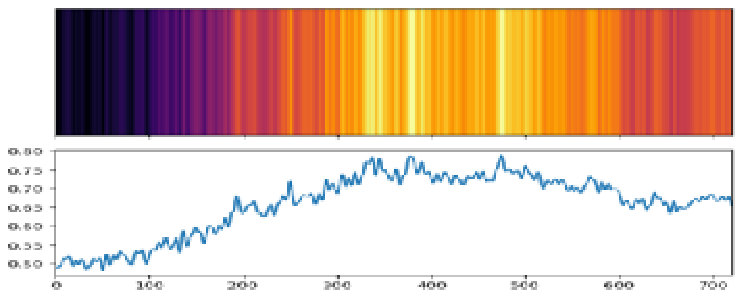}
        };\end{tikzpicture} & \multirow{8}{0.4cm}{\begin{tikzpicture}\node[inner sep=0pt]{\includegraphics[width=0.2cm,height=8.5cm,bb=0 0 14 301]{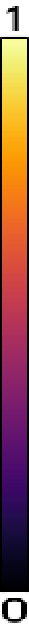}};\end{tikzpicture}} \\
        FiLM & 0.26652 & \begin{tikzpicture} \node[inner sep=0pt] {
            \includegraphics[width=5cm,height=1cm]{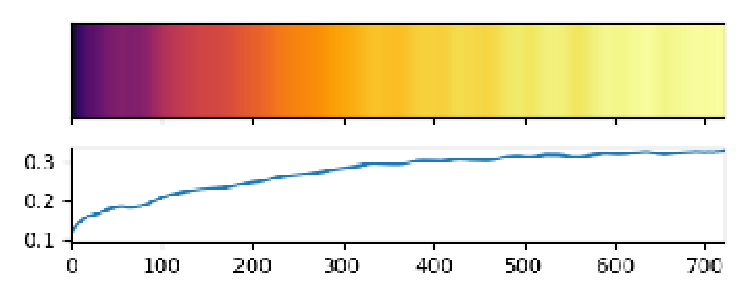}
        };\end{tikzpicture} & 0.35619 & \begin{tikzpicture} \node[inner sep=0pt] {
            \includegraphics[width=5cm,height=1cm]{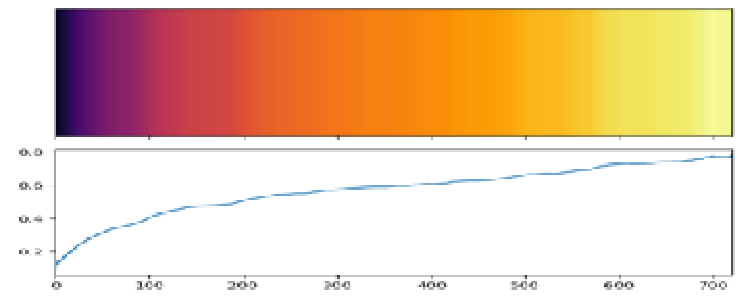}
        };\end{tikzpicture} & \\
        PatchTST & 0.57651 & \begin{tikzpicture} \node[inner sep=0pt] {
            \includegraphics[width=5cm,height=1cm]{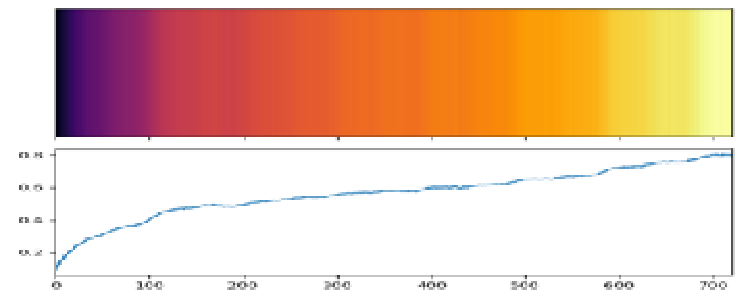}
        };\end{tikzpicture} & 0.35941 & \begin{tikzpicture} \node[inner sep=0pt] {
            \includegraphics[width=5cm,height=1cm]{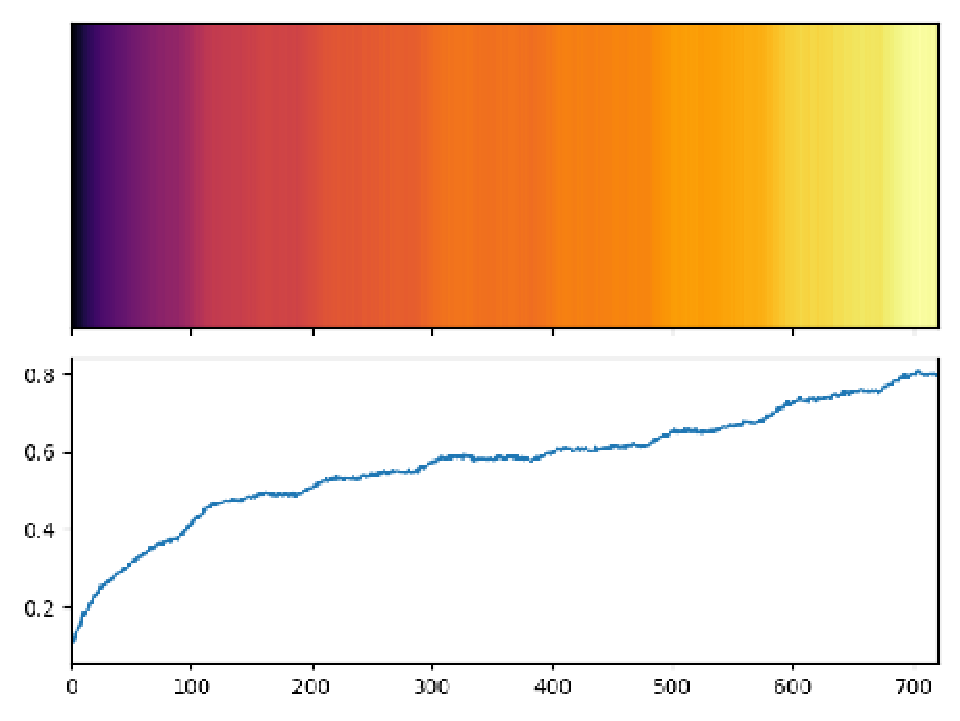}
        };\end{tikzpicture} & \\
        xLSTM & 0.32941 & \begin{tikzpicture} \node[inner sep=0pt] {
            \includegraphics[width=5cm,height=1cm]{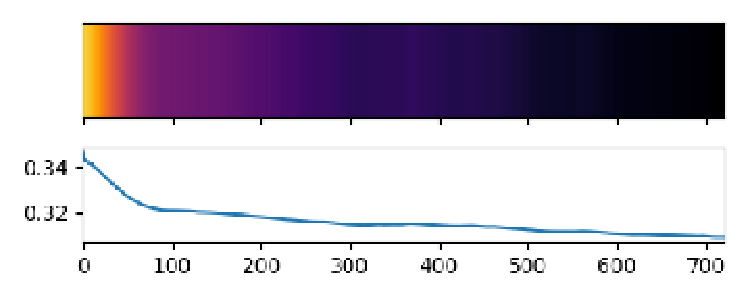}
        };\end{tikzpicture} & 0.31646 & \begin{tikzpicture} \node[inner sep=0pt] {
            \includegraphics[width=5cm,height=1cm]{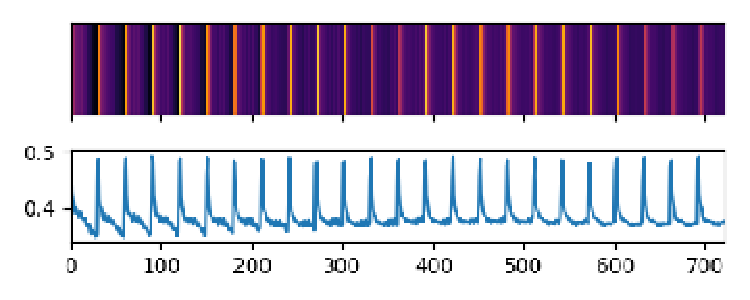}
        };\end{tikzpicture} & \\
        Triformer & 0.36018 & \begin{tikzpicture} \node[inner sep=0pt] {
            \includegraphics[width=5cm,height=1cm]{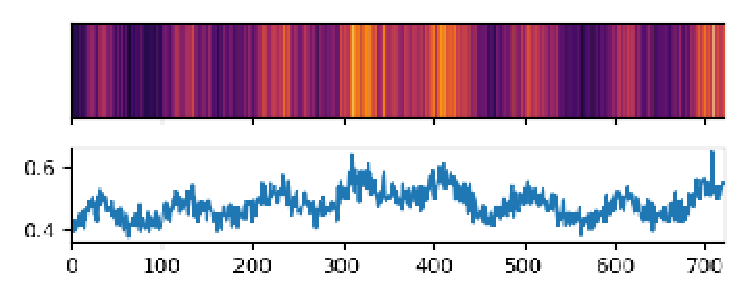}
        };\end{tikzpicture} & 0.25790 & \begin{tikzpicture} \node[inner sep=0pt] {
            \includegraphics[width=5cm,height=1cm]{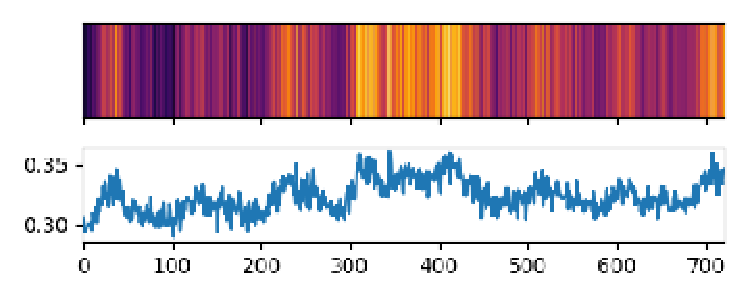}
        };\end{tikzpicture} & \\
        FEDformer & 0.33840 & \begin{tikzpicture} \node[inner sep=0pt] {
            \includegraphics[width=5cm,height=1cm]{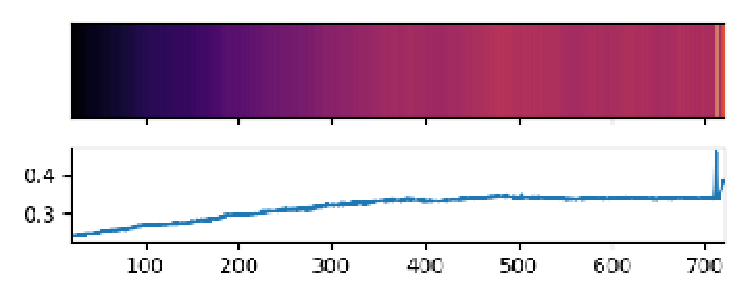}
        };\end{tikzpicture} & 0.21476 & \begin{tikzpicture} \node[inner sep=0pt] {
            \includegraphics[width=5cm,height=1cm]{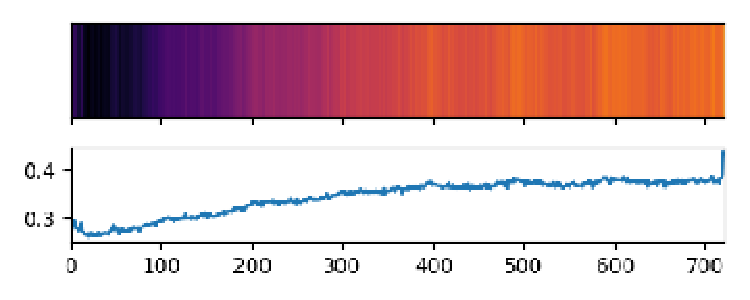}
        };\end{tikzpicture} & \\
        N-Linear & 0.28170 & \begin{tikzpicture} \node[inner sep=0pt] {
            \includegraphics[width=5cm,height=1cm]{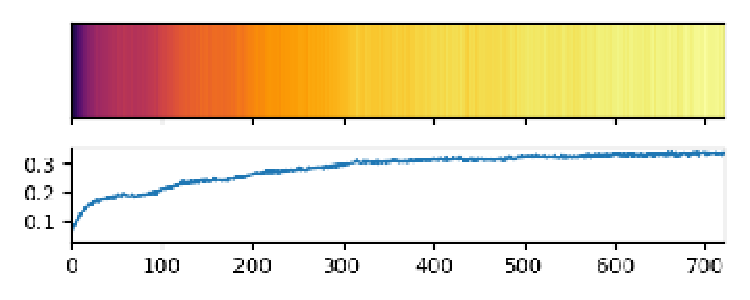}
        };\end{tikzpicture} & 0.21101 & \begin{tikzpicture} \node[inner sep=0pt] {
            \includegraphics[width=5cm,height=1cm]{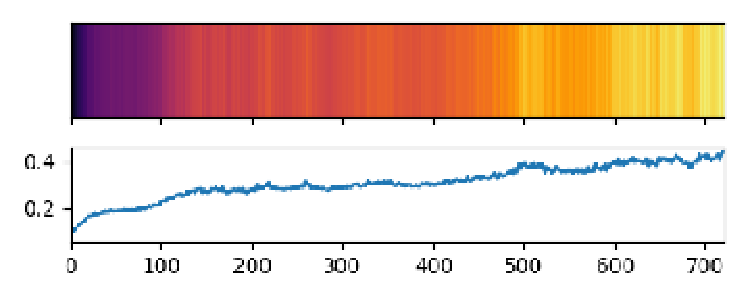}
        };\end{tikzpicture} & \\
        NHITS & 0.29218 & \begin{tikzpicture} \node[inner sep=0pt] {
            \includegraphics[width=5cm,height=1cm]{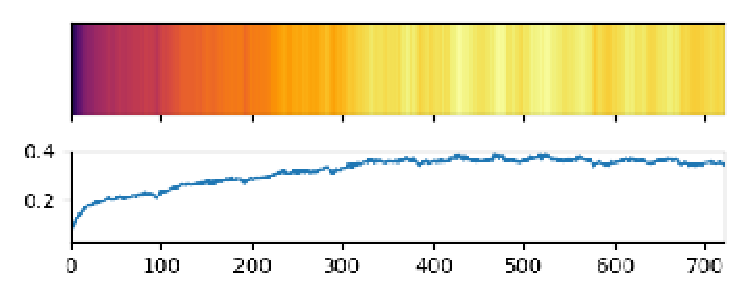}
            };\end{tikzpicture} & 0.20643 & \begin{tikzpicture}
            \node[inner sep=0pt] {
            \includegraphics[width=5cm,height=1cm]{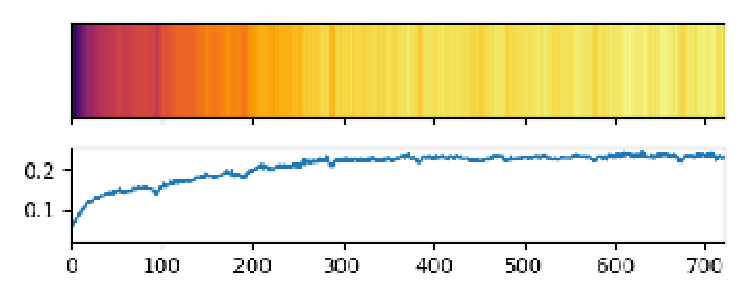}
            };\end{tikzpicture} \\
        \bottomrule
    \end{tabular}}
    \caption{Using the trained univariate HUFL and multivariate ETTm2 models and its test set, predictions are generated over a horizon of length 720. The MSE scores of every predicted horizon are averaged across the test set per-timestep. The performance differences characterize the LHF problem.}
    \label{table4}
\end{table}
\subsection{Mathematics of Deep Learning Models for Series Generation}
\emph{PatchTST} in the data space, the following in the feature space, 1) the convolutional windows in \emph{LogSparse transformer}, 2) the time-delay aggregation in \emph{Autoformer}, 3) the convolutional pyramid attention in \emph{Pyraformer} and 4) \emph{Triformer} in both the data and feature spaces, and 5) pooling in \emph{NHITS}, show that the use of patching helps increase the length of the horizon of forecasts. \emph{PatchTST} further shows that splitting the features across attention heads alongwith the patching helps improve the performance of multi-variate forecasting over other self-attention models and is slightly better than \emph{D-Linear}. This channel independence follows the same observations that have led to the use of sparsity, perceptron architectures and attention to prevent overfitting because of the number of model parameters. \emph{NBEATS} and \emph{NHITS} predict the series one series at a time and learn representations for each of them in a shared feature space, whereas the others predict the series altogether. While the dimensionality of the input is small enough for the differences not to affect performance, gradient descent finds fundamentally different landscapes in these methods. There seem to be variations in performances between \emph{NHITS} and \emph{N-Linear} multivariate forecasting models (from table \ref{table3}) even though the error averages are similar, highlighting this difference. The former class of models impose that the parameters average between the series, while the latter acknowledges the existence of a shared space that can be loosely likened to co-integrated variables in the feature space. Compared to sequences, time series data use smaller datasets and work with 1-dimensional input feature(s) as opposed to language embeddings and thus self-attention models with softmax operations and multi-layer perceptron layers with ReLU activations seem to perform just as well as either. \emph{FEDformer}, \emph{Autoformer}, \emph{NBEATS}, and \emph{FiLM} use the frequency domain of the time series data as a part of the layers or input. Patching hasn't yet been implemented using existing multi-layer perceptron models, although the pooling and hierarchical interpolation in \emph{NHITS} could be considered as analogous. Low-rank matrix approximation is effective in reducing the number of parameters in \emph{FiLM} and factorization is used to create layer-agnostic reduction in number of parameters by maintaining the left and right matrices common in \emph{Triformer}. A highway connection over the exogenous variables improves contextual learning in \emph{TiDE}. The use of the attention mask to ensure causality in self-attention has been used to define the pyramidal attention in \emph{Pyraformer}, after the deep strided convolution layers. The dependence between the choice of layers and the type of multivariate forecasting seems to remain a question because while \emph{NHITS} performs better in the single-input setting, \emph{N-Linear} performs better in the multiple-input multivariate setting, with minimal performance differences in \emph{NHITS} because of the dataset size reduction.  
\subsection{Number of Parameters}
Self-attention models that use the frequency domain are able to perform better on univariate data and this questions the claims in (\cite{zeng2023transformers}). In general, the frequency domain transformations in the feature space enabled the use of more parameters without being overfit and mostly performed better with the exception of \emph{N-Linear} in the multivariate forecasting experiments. The State Space Models (SSMs) family has also been used for long horizon forecasting, has number of parameters around 5 times smaller than \emph{N-Linear} while performing similar, but hasn't been included in this review. The SSM model \emph{SpaceTime} (\cite{zhang2023effectively}) uses companion SSMs to forecast longer horizons better, with experiments involving horizons longer than 720 showing promising results and they show results slightly better than the models in this review. The improvements to the scores reported in the cited work highlight the role dataset size could play in time series forecasting and encourages the collection of data over longer periods of time. 

\bibliographystyle{elsarticle-harv}
\bibliography{ref}

\end{document}